\documentclass[lettersize,journal]{IEEEtran}
\usepackage{amsmath,amsfonts}
\usepackage{algorithmic}
\usepackage{algorithm}
\usepackage{array}
\usepackage[caption=false,font=normalsize,labelfont=sf,textfont=sf]{subfig}
\usepackage{textcomp}
\usepackage{stfloats}
\usepackage{url}
\usepackage{verbatim}
\usepackage{graphicx}
\usepackage{cite}
\usepackage{xcolor}
\usepackage{makecell}
\usepackage{multirow}
\usepackage{tabularx}
\usepackage{booktabs}
\usepackage{hyperref}

\definecolor{codebgcolor}{HTML}{EDEDED}

\begin{document}

\title{QFFN-BERT: An Empirical Study of Depth, Performance, and Data Efficiency in Hybrid Quantum-Classical Transformers}

\author{Pilsung Kang
\thanks{This work was supported by the National Research Foundation of Korea (NRF) grant funded by the Korea government (MSIT), grant number RS-2023-00244605.}
\thanks{Pilsung Kang is with the Department of Software Science, Dankook University, Yongin 16890, South Korea (e-mail: pilsungk@dankook.ac.kr)}
}

\markboth{Journal of \LaTeX\ Class Files,~Vol.~14, No.~8, August~2021}%
{Shell \MakeLowercase{\textit{et al.}}: A Sample Article Using IEEEtran.cls for IEEE Journals}


\maketitle

\begin{abstract}
Parameterized quantum circuits (PQCs) have recently emerged as promising components for enhancing the expressibility of neural architectures. In this work, we introduce QFFN-BERT, a hybrid quantum-classical transformer where the feedforward network (FFN) modules of a compact BERT variant are replaced by PQC-based layers. This design is motivated by the dominant parameter contribution of FFNs, which account for approximately two-thirds of the parameters within standard Transformer encoder blocks.  While prior studies have primarily integrated PQCs into self-attention modules, our work focuses on the FFN and systematically investigates the trade-offs between PQC depth, expressibility, and trainability. Our final PQC architecture incorporates a residual connection, both $R_Y$ and $R_Z$ rotations, and an alternating entanglement strategy to ensure stable training and high expressibility.  Our experiments, conducted on a classical simulator, on the SST-2 and DBpedia benchmarks demonstrate two key findings. First, a carefully configured QFFN-BERT achieves up to 102.0\% of the baseline accuracy, surpassing its classical counterpart in a full-data setting while reducing FFN-specific parameters by over 99\%.  Second, our model exhibits a consistent and competitive edge in few-shot learning scenarios, confirming its potential for superior data efficiency.  These results, supported by an ablation study on a non-optimized PQC that failed to learn, confirm that PQCs can serve as powerful and parameter-efficient alternatives to classical FFNs when co-designed with foundational deep learning principles.
\end{abstract}


\section{Introduction}
\label{s:intro}
Large language models (LLMs) such as BERT~\cite{devlin:2019:bert} and GPT~\cite{radford:2018:gpt1} have revolutionized natural language processing (NLP) by achieving state-of-the-art performance on tasks like sentiment classification, question answering, and textual inference. These models are predominantly based on the Transformer architecture~\cite{vaswani:2017:nips}, which utilizes large FFNs in each encoder layer. In the widely-used BERT-base model, where the hidden size is 768 and the FFN intermediate size is 3072, these FFNs account for a substantial portion of the model's parameters. This specific architectural characteristic has been identified as a key source of inefficiency in subsequent works such as ALBERT~\cite{lan:2020:iclr}, making FFNs a prime target for optimization and innovation. Despite their success, such models face well-documented challenges in terms of inference cost, memory footprint, and energy consumption—factors that complicate deployment in resource-constrained or real-time environments.

Efforts to mitigate these costs have primarily focused on compression and parameter reduction within classical architectures, including quantization~\cite{shen:2020:aaai}, knowledge distillation~\cite{sanh:2020:distilbert,sun:2019:acl,wang:2020:nips}, and low-rank or parameter-efficient adaptation~\cite{hu2021lora,houlsby:2019:icml}. However, these approaches are typically confined to classical structures and often yield trade-offs between performance and compactness. 

The emerging field of quantum machine learning (QML) offers a potential alternative. Hybrid quantum-classical approaches, where a PQC is optimized by a classical computer, have become a leading paradigm for near-term quantum computing, first established in foundational works such as the Variational Quantum Eigensolver (VQE)~\cite{peruzzo:2014:vqe}. Our work builds on this paradigm by embedding a PQC into a classical Transformer to leverage quantum state spaces for expressive computation. Notably, while PQCs have been proposed for other architectural components, their role in FFNs---a major architectural element---remains underexplored. Recent studies, for example, have focused on integrating PQCs into self-attention modules, as seen in quantum self-attention neural networks~\cite{li:2024:qsann}, or into vision models like quantum vision transformers~\cite{cherrat:2024:qvt}. In most of these hybrid models, the FFN typically remains classical or receives only minor quantum enhancements. Given the structural suitability of FFNs for low-dimensional, position-wise PQC implementations, this gap motivates our targeted exploration into fully quantum-enhanced FFNs.

In this work, we propose a new hybrid architecture by replacing the FFN modules in a compact BERT model with PQC-based layers. Our implementation uses Qiskit~\cite{javadiabhari:2024:qiskit} for the quantum components, integrated into a PyTorch~\cite{paszke:2019:nips} framework via \texttt{TorchConnector}. Our motivation stems from the structural alignment between FFN layers and NISQ~\cite{preskill:2018:nisq} hardware constraints: unlike attention modules, FFNs operate independently across token positions and are amenable to reduced-dimensional quantum embedding. By targeting FFNs—rather than more commonly studied attention blocks—we aim to examine whether quantum circuits can improve parameter efficiency and generalization without sacrificing task performance. Although the full replacement of FFN structures with quantum circuits is not currently feasible under the limitations of NISQ-era hardware—particularly due to constraints on input dimensionality and circuit depth—our design intentionally restricts quantum integration to low-dimensional modules that structurally align with FFN behavior.

We systematically vary the PQC depth to investigate its impact on model expressibility, generalization, and optimization dynamics. This depth-scaling analysis allows us to empirically assess the trade-off between the enhanced representational power of deeper circuits and the trainability challenges predicted by barren plateau phenomena. Rather than proposing a production-ready solution, our study focuses on characterizing the advantages and limitations of replacing classical FFNs with these quantum counterparts in a simulated environment.

We consider FFNs a particularly suitable target for quantum integration for several reasons. First, FFNs account for the majority of parameters and nonlinear transformations within Transformer blocks, and are therefore natural candidates for structural substitution aimed at improving performance and efficiency. Second, the position-wise nature of FFNs means the same transformation is applied independently to each token. This structure avoids complex inter-token dependencies and makes the entire FFN block conceptually replaceable by a single, small PQC, greatly reducing the quantum resource requirements compared to replacing other modules like attention. Third, FFNs contribute most of the model's expressive power through ReLU-based transformations, a role that PQCs can potentially fulfill through unitary parameterization and entanglement. These structural and functional parallels motivate our investigation into PQC-enhanced FFNs as a testbed for quantum-enhanced representation learning within practical Transformer models.

The central hypothesis of this work is that quantum-enhanced FFNs can maintain or exceed competitive accuracy while offering significant benefits in performance, generalization, and parameter efficiency. In particular, we hypothesize that PQCs can provide competitive expressibility compared to classical feedforward networks while requiring significantly fewer trainable parameters \textit{at the module level}. This offers an architecturally efficient and highly expressive alternative for transformer-based architectures. Although current quantum backends do not offer inference speed advantages, we argue that architectural gains in expressivity and data efficiency remain feasible within simulated or limited-qubit environments. Furthermore, by systematically varying the depth of PQC layers, we aim to empirically assess the trade-offs between improved performance and potential optimization challenges. This depth scaling analysis provides insights into the practical limitations and opportunities for quantum-enhanced architectures.

The contributions of this paper are summarized as follows:
\begin{itemize}
    \item \textbf{A Novel and Effective Hybrid Architecture:} We propose and validate QFFN-BERT, a new architecture demonstrating that a PQC block can successfully replace the FFN in a Transformer, achieving performance that surpasses a strong classical baseline.

    \item \textbf{Empirical Analysis of the Expressibility-Trainability Trade-off:} We provide a systematic analysis of PQC depth, empirically demonstrating the trade-off between increased expressibility at lower depths and the onset of optimization challenges (i.e., barren plateaus) at greater depths, identifying an empirical sweet spot for performance.

    \item \textbf{Demonstration of Superior Data Efficiency:} We show that QFFN-BERT exhibits a consistent and competitive performance edge in few-shot learning scenarios, highlighting the potential of quantum-enhanced models for data-efficient NLP.

    \item \textbf{Validation of Critical Co-Design Principles:} Through a detailed ablation study, we prove that the success of deep hybrid models requires a deliberate co-design, highlighting the indispensable role of foundational deep learning principles (e.g., residual connections) and sophisticated PQC designs that enhance trainability.
\end{itemize}

The remainder of this paper is structured as follows. Section~\ref{s:related} reviews existing studies on model compression techniques, hybrid quantum-classical architectures, and theoretical insights into quantum expressibility and optimization challenges. Section~\ref{s:setup} describes the evaluation setup, including the datasets, experimental configurations, and PQC depth scaling methodology. Section~\ref{s:eval} presents the evaluation results, analyzing the impact of PQC depth on model accuracy, generalization, and parameter efficiency. Section~\ref{s:ablation} provides an in-depth ablation study, highlighting the critical impact of our architectural optimizations by contrast with a baseline quantum design. Finally, Section~\ref{s:conc} concludes the paper with a summary of our contributions and directions for future work.

\section{Related Work}
\label{s:related}

\subsection{Feedforward Network Compression in Transformers}

The FFN component of Transformer architectures has been a focal point for model compression efforts. MobileBERT~\cite{sun:2020:mobilebert} introduces a bottleneck structure within FFNs to reduce model size and latency, achieving a 4.3x smaller and 5.5x faster model than BERT\textsubscript{BASE} while maintaining competitive performance. Similarly, TinyBERT~\cite{jiao:2020:tinybert} employs knowledge distillation to compress both the attention and FFN layers, resulting in a model that is 7.5x smaller and 9.4x faster than BERT\textsubscript{BASE}. LoRA~\cite{hu2021lora} proposes low-rank adaptations within Transformer layers, including FFNs, to enable efficient fine-tuning with reduced parameter updates. While these methods effectively compress FFNs within classical frameworks, they do not explore quantum enhancements or the integration of PQCs.

\subsection{Hybrid Quantum-Classical Architectures}

Hybrid quantum-classical models have emerged as a promising avenue for leveraging quantum computing in machine learning. Havl\'{\i}\v{c}ek et al.~\cite{havlicek:2019:nature} demonstrate the use of quantum-enhanced feature spaces for classification tasks, employing PQCs to map classical data into high-dimensional quantum Hilbert spaces. Farhi and Neven~\cite{farhi2018qnn} 
propose a quantum neural network (QNN) framework designed for near-term quantum processors, demonstrating through simulation that QNNs can classify both classical and quantum data using PQCs and supervised learning methods. Quanvolutional neural networks~\cite{henderson:2020:quanvol} is a hybrid architecture that incorporates quantum circuit-based transformations into classical convolutional neural networks to extract potentially useful features from image data. These studies primarily focus on integrating quantum components into attention mechanisms or convolutional layers, leaving the FFN structure within Transformers largely unexplored in the context of quantum integration.

\subsection{Quantum Circuits in Natural Language Processing}

The application of quantum circuits to natural language processing (NLP) tasks has been investigated in various contexts. Coecke et al.~\cite{coecke2020foundations} provide the conceptual and mathematical foundations for quantum natural language processing (QNLP), introducing a framework that encodes grammatical and semantic structures into quantum circuits using categorical models and variational quantum techniques, paving the way for practical implementations on near-term quantum hardware.
Di Sipio et al.~\cite{sipio:2022:icassp} implement and evaluate a quantum-enhanced LSTM for part-of-speech tagging and propose a quantum-enhanced Transformer for sentiment analysis. Their preliminary results show that quantum-classical hybrid models can achieve comparable performance to classical counterparts while using significantly fewer parameters, although training costs remain high.

\subsection{Barren Plateau Phenomenon in Quantum Neural Networks}

The barren plateau phenomenon, characterized by vanishing gradients in the optimization landscape of PQCs, poses a significant challenge for training quantum neural networks. Cerezo et al.~\cite{cerezo:2021:nature} analyze how cost function choices influence the presence of barren plateaus, providing strategies to mitigate this issue in shallow circuits. Pesah et al.~\cite{pesah:2021:aps} demonstrate that certain architectures, such as quantum convolutional neural networks, can inherently avoid barren plateaus due to their structural properties. These insights are crucial for designing trainable quantum-enhanced FFNs, as they inform the selection of PQC architectures that balance expressibility with trainability.

\subsection{Summary of Key Differences}

While prior research has extensively explored FFN compression within classical frameworks and the integration of quantum components into various neural network architectures, the specific application of PQCs to FFN layers in pre-trained Transformers remains underinvestigated. Our work addresses this gap by replacing the FFN module in each encoder layer of a lightweight BERT model with PQCs, implemented via Qiskit and integrated into PyTorch using \texttt{TorchConnector}. We systematically vary the depth of PQC layers to assess their impact on model expressibility, generalization behavior, and optimization dynamics, particularly in the context of the barren plateau phenomenon. To our knowledge, this is the first study to explore the integration of PQC-based FFNs within a practical, pre-trained Transformer backbone, providing empirical insights into the feasibility and benefits of quantum-enhanced FFNs in practical NLP applications.

\section{Experimental Setup}
\label{s:setup}

To evaluate the feasibility and trade-offs of integrating PQCs into the FFNs of Transformer architectures, we conduct experiments across varying quantum circuit depths on two representative NLP benchmarks. Our evaluation is centered around a core set of benchmark metrics designed to systematically assess the impact of PQC depth on expressibility, generalization, and optimization dynamics.

To systematically compare the classical and hybrid FFN designs, we evaluate a core set of metrics including validation accuracy, parameter efficiency, and few-shot performance. By analyzing these metrics, particularly the accuracy-per-parameter ratio and the train-validation gap, we quantify the practical trade-offs of our proposed architecture. This section details the complete experimental setup used for this investigation, covering the development environment, datasets, model variants, PQC design, and training protocols.

\subsection{Development Environment and Software Frameworks}

Table~\ref{tab:env} summarizes the hardware and software configurations used in this study. The system was equipped with two Intel Xeon Bronze 3106 CPUs operating at 1.70GHz, providing a total of 16 physical cores without hyper-threading, and 128 GB of RAM. The operating system was Ubuntu 22.04 LTS. The model implementations were based on PyTorch 2.7.0, while dataset handling and pre-trained model loading utilized HuggingFace Transformers 4.51.3 and HuggingFace Datasets 3.5.0. Quantum components were developed using Qiskit 1.0.2, and integration with classical models was achieved through the \texttt{TorchConnector} module from Qiskit Machine Learning. Python 3.10.12 served as the development language across all experiments.

All experiments were executed in CPU-only mode, as GPU acceleration did not yield significant performance improvements due to the inherently CPU-bound nature of quantum circuit simulation in Qiskit.  

\begin{table}[hbtp]
\centering
\caption{Summary of Hardware and Software Environment}
\label{tab:env}
\begin{tabular*}{0.9\columnwidth}{@{\extracolsep{\fill}}ll@{}}
\toprule
\textbf{Component} & \textbf{Specification} \\
\midrule
CPU & \makecell[l]{2 × Intel Xeon Bronze 3106 \\ @ 1.70GHz (16 cores total)} \\
RAM & 128 GB \\
Operating System & Ubuntu 22.04 LTS \\
Python Version & 3.10.12 \\
PyTorch Version & 2.7.0 \\
HuggingFace Transformers & 4.51.3 \\
HuggingFace Datasets & 3.5.0 \\
Qiskit Version & 1.0.2 \\
Other Libraries & \makecell[l]{TorchConnector \\ (from Qiskit Machine Learning)} \\
\bottomrule
\end{tabular*}
\end{table}

\subsection{Evaluated Datasets}
\label{ss:datasets}

To evaluate the proposed models under varying task complexity and input characteristics, we adopt two representative NLP datasets: SST-2~\cite{socher:2013:recursive} and DBpedia~\cite{hasibi:2017:DVT,muennighoff:2022:mteb}. SST-2 is a binary sentiment classification task involving short movie review sentences, while DBpedia is a 14-way topic classification benchmark composed of structured Wikipedia abstracts. The contrast between these datasets enables a controlled analysis of model behavior in both low-complexity and high-complexity regimes.

All textual inputs are tokenized using the TinyBERT tokenizer with a maximum input length of 128 tokens. For SST-2, the average number of tokens per input is 13.32, reflecting the brevity and simplicity of the sentiment classification task. In contrast, DBpedia exhibits an average of 63.46 tokens per input after tokenization, introducing significantly greater sequence-level complexity. This discrepancy imposes different computational loads on the PQC-based FFN modules, making DBpedia particularly suitable for examining depth-expressibility trade-offs in longer-context settings.  A summary of the dataset statistics is provided in Table~\ref{tab:datasets}.

No additional preprocessing (e.g., lowercasing, stopword removal) was applied to either dataset. Standard train/test splits from the HuggingFace Datasets library were used.

\begin{table}[hbtp]
\centering
\caption{Summary of the NLP Datasets used for evaluation}
\label{tab:datasets}
\begin{tabularx}{\columnwidth}{@{}lXX@{}}
\toprule
\textbf{Property} & \textbf{SST-2} & \textbf{DBpedia} \\
\midrule
Description & Stanford Sentiment Treebank (movie reviews) & Wikipedia-based Topics \\
\addlinespace 
Task Type & Sentiment Classification & Topic Classification \\
\addlinespace
Classes & Positive, Negative & 14 Named Topics \\
\addlinespace
Training Size & 67,349 & 560,000 \\
\addlinespace
Validation / \\Test Size & 872 / 1,821 & --- / 70,000 \\
\addlinespace
Avg. Token Length & 13.32 & 63.46 \\
\bottomrule
\end{tabularx}
\end{table}

\subsection{Model Architecture Variants}
\label{ss:model}

The baseline architecture used in this study is a lightweight BERT model from HuggingFace, \texttt{prajjwal1/bert-tiny}~\cite{turc:2019:wellread,bhargava:2021:generalization}, which consists of 2 transformer layers and a 128-dimensional hidden size. Note that this model is structurally different from ``TinyBERT''~\cite{jiao:2020:tinybert}, which is a distilled variant of BERT. To avoid confusion, we refer to our baseline model simply as ``\texttt{bert-tiny}'' throughout this paper.

To ensure reproducibility and efficient experimentation across multiple PQC-based FFN configurations, we adopt this compact \texttt{bert-tiny} model as the base architecture instead of standard BERT. This choice offers two practical advantages: (1) it significantly reduces computational cost, allowing systematic depth scaling and few-shot evaluations under limited resources; and (2) it preserves the essential structure of Transformer-based encoders, ensuring that the impact of PQC-based FFN replacement can be meaningfully assessed. As our goal is to analyze architectural trade-offs rather than absolute SOTA performance, the \texttt{bert-tiny} model provides a balanced and efficient experimental platform.

In our hybrid quantum-classical variants, we replace each classical FFN module in this \texttt{bert-tiny} model with a \textbf{Quantum Feedforward Network (QFFN)}, a block composed of PQCs. The resulting model is referred to as \textbf{QFFN-BERT}. The QFFN block is composed of three stages: (1) a classical linear projection that maps the hidden representation to a 4-dimensional input suitable for quantum processing, (2) a multi-layer PQC module that processes this input using alternating $R_Z$ and $R_Y$ rotations interleaved with engtangling gates (CNOT and CZ), and (3) a final classical projection that restores the hidden dimensionality.

Each QFFN variant differs in the number of stacked PQC layers per FFN replacement module, allowing us to analyze how increased circuit depth affects expressibility, generalization, and trainability. The following depth settings are evaluated:

\begin{itemize}
    \item \textbf{QFFN-BERT (1L)}: One PQC layer per FFN block
    \item \textbf{QFFN-BERT (2L)}: Two PQC layers per FFN block
    \item \textbf{QFFN-BERT (4L)}: Four PQC layers per FFN block
    \item \textbf{QFFN-BERT (8L)}: Eight PQC layers per FFN block
\end{itemize}

While all four depth settings were evaluated on the SST-2 dataset, the 8-layer configuration was omitted for the significantly larger DBpedia dataset due to its computationally prohibitive runtime under simulation.

All other architectural components, including embeddings and self-attention modules, are left unchanged. The number of qubits in our quantum circuits is fixed at four across all configurations. A key aspect of our design is that the PQC is applied only to the [CLS] token representation. This design choice was made for two primary reasons. First, for the sentence-level classification tasks used in our benchmarks, the final representation of the [CLS] token is the most critical component for prediction. Second, given the significant computational cost of quantum circuit simulation, applying the PQC to every token in the sequence was computationally prohibitive for our systematic experiments. While this represents a deviation from the standard position-wise FFN, it allows for a focused and feasible investigation into the PQC's ability to act as a powerful non-linear transformation block for sentence-level tasks. This overall modular replacement strategy enables controlled studies on the performance–efficiency–trainability trade-offs introduced by PQC-based FFN substitution.

\begin{figure}[!htb]
    \centering
    \includegraphics[width=0.47\textwidth]{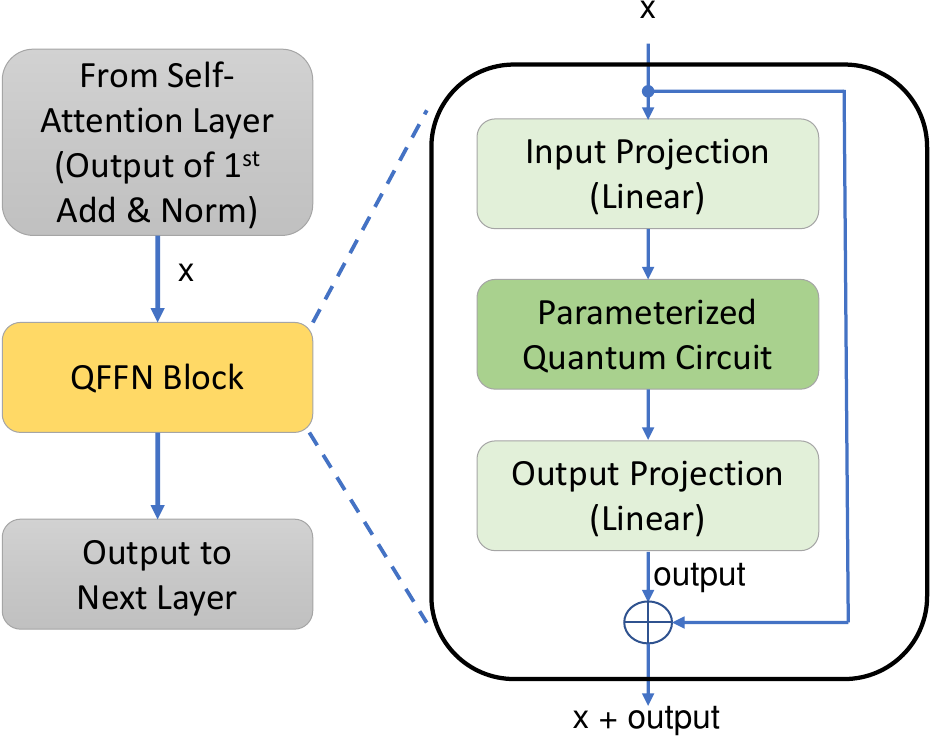}
    \caption{The architecture of a single, modified Transformer encoder layer in QFFN-BERT. The output from the self-attention sub-layer enters our proposed QFFN block, which replaces the standard FFN. The QFFN block's internal structure consists of input/output linear projections and a PQC, and crucially incorporates an internal residual connection that adds the block's input $x$ to the transformed output.}

    \label{fig:qffn-bert}
\end{figure}

Fig.~\ref{fig:qffn-bert} illustrates the overall architecture of QFFN-BERT, highlighting the integration of deep PQC blocks in place of classical FFNs.

\subsection{PQC Design}

We adopt a deep PQC structure for replacing the FFN modules within the \texttt{bert-tiny} backbone.  Each PQC block consists of multiple stacked layers, where each layer comprises a trainable quantum transformation that acts on a 4-dimensional projected input.

The internal structure of each PQC layer consists of the following stages:

\begin{itemize}
    \item \textbf{Single Data Re-encoding:} At the first layer only, each qubit is initialized using an $R_Y(x)$ rotation, where $x$ is the projected classical feature. This stage embeds classical information into the quantum amplitude space. Unlike shallow PQC designs that reapply data encoding at each layer, we perform this operation once to reduce circuit depth and emphasize quantum evolution.

    \item \textbf{Layer-wise Entanglement:} We alternate the entanglement pattern across layers for improved expressibility. Even-numbered layers apply a circular chain of \texttt{CNOT} gates (q$_0$ $\rightarrow$ q$_1$ $\rightarrow$ q$_2$ $\rightarrow$ q$_3$ $\rightarrow$ q$_0$), while odd-numbered layers apply \texttt{CZ} gates between non-adjacent qubit pairs (q$_0$–q$_2$, q$_1$–q$_3$). This alternating topology enriches nonlocal correlations and improves gradient flow.

    \item \textbf{Trainable Rotations:} Each layer includes a trainable $R_Z(\theta)$ rotation followed by a trainable $R_Y(\theta)$ rotation on every qubit. This RZ–RY combination increases the parameter space dimensionality and directional flexibility, enabling the PQC to approximate a broader class of nonlinear functions.
\end{itemize}

\begin{figure*}[!htb]
    \centering
    \includegraphics[width=0.9\textwidth]{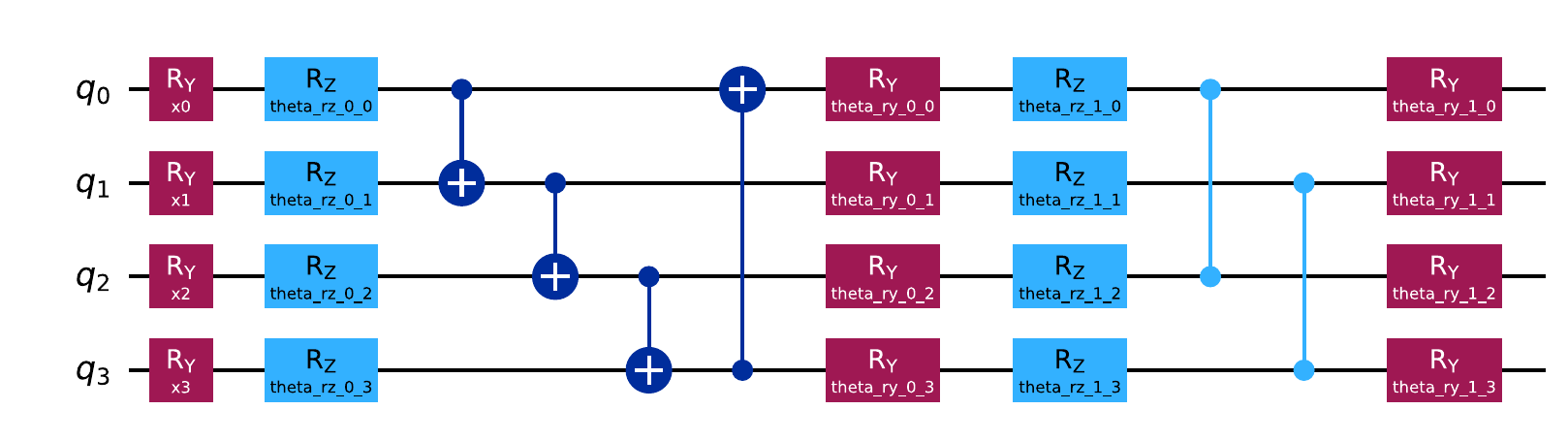}
    \caption{Structure of two stacked PQC layers: classical inputs $x_i$ are embedded at layer 1 via $R_Y(x_i)$ rotations. Each layer applies an $R_Z$–$R_Y$ trainable rotation pair per qubit and alternates entanglement patterns (CNOT, CZ) across layers.}
    \label{fig:pqc_circuit}
\end{figure*}

This sequence---data encoding (first layer only), entanglement (alternating CZ/CNOT), and trainable rotations (RZ–RY)---forms a single PQC layer.  Multiple such layers are stacked to form deep PQC blocks with $\{1, 2, 4, 8\}$ layers evaluated across experiments. The full PQC block is simulated using Qiskit's \texttt{EstimatorQNN} interface and integrated into the \texttt{bert-tiny} encoder via residual connections, replacing the original FFN module. Fig.~\ref{fig:pqc_circuit} illustrates the two-layer variant used in this study. All models share the same 4-qubit configuration for fair comparison, and no circuit-level optimizations were applied.

While structurally simple and expressively designed, the PQC block constitutes a major bottleneck in wall-clock runtime. Due to the CPU-based simulation of quantum circuits in Qiskit, PQC evaluation becomes significantly slower than the surrounding GPU-accelerated components of \texttt{bert-tiny}. This computational gap is particularly evident at higher depth configurations, where the PQC—despite being applied only to the [CLS] token—accounts for more than 90\% of the forward pass time per batch.

This bottleneck scales approximately linearly with PQC depth and highlights the need for optimization-aware PQC design not only from an accuracy or generalization standpoint, but also for ensuring computational efficiency. Our findings reinforce the broader insight that variational quantum modules must be co-developed with their classical execution environments in mind, especially in simulation-based research workflows prior to quantum hardware deployment.

\subsection{Training and Fine-Tuning Settings}

All models were fine-tuned on the SST-2 and DBpedia datasets using a consistent training protocol to ensure a fair comparison. We utilized the HuggingFace Transformers library~\cite{wolf:2020:huggingface} for model loading and tokenization, with PyTorch serving as the backend for our custom training loops. 

The training was optimized using cross-entropy loss for a maximum of 5 epochs, with a fixed random seed of 42 for reproducibility. Key hyperparameters for the classical optimization process, such as the learning rate and batch size, are summarized in Table~\ref{tab:hyperparams}. For the quantum components, the trainable PQC parameters were initialized from a uniform distribution between $-\pi$ and $\pi$. Gradients for these parameters were calculated using the standard parameter-shift rule. No advanced techniques such as learning rate schedulers, data augmentation, or early stopping were employed to isolate the impact of the architectural changes.

All experiments, including few-shot scenarios created by subsampling the training data, were conducted on a CPU-only infrastructure due to the simulation-based nature of our quantum components.

\begin{table}[hbtp]
\centering
\caption{Key Hyperparameters for Model Training}
\label{tab:hyperparams}
\begin{tabular*}{0.9\columnwidth}{@{\extracolsep{\fill}}ll@{}}
\toprule
\textbf{Hyperparameter} & \textbf{Value} \\
\midrule
Pre-trained Model & \texttt{prajjwal1/bert-tiny} \\
Optimizer & Adam \\
Learning Rate & 5E-4 \\
Max Epochs & 5 \\
Effective Batch Size & 32 \\
Max Sequence Length & 128 tokens \\
Random Seed & 42 \\
Loss Function & Cross-Entropy \\
PQC Param. Initialization & Uniform $(-\pi, \pi)$ \\
Quantum Gradient Method & Parameter-shift rule \\
\bottomrule
\end{tabular*}
\end{table}

\subsection{Evaluation Metrics}

We adopt the following metrics to systematically evaluate models trained with PQC-enhanced and classical FFN modules:

\begin{itemize}
    \item \textbf{Validation Accuracy:} The proportion of correctly predicted examples on the validation set.
    
    \item \textbf{Train-Validation Gap:} The difference between training accuracy and validation accuracy, measuring generalization behavior.
    
    \item \textbf{Accuracy per Parameter:} The ratio of validation accuracy to the total number of trainable parameters, assessing compactness and efficiency.
    
    \item \textbf{Few-shot Performance:} Validation accuracy measured when training with 10\% or 20\% of the full training data, analyzing robustness in low-data regimes.
\end{itemize}

All metrics are computed directly based on model outputs without additional calibration, ensembling, or post-processing.

\section{Evaluation Results}
\label{s:eval}

\subsection{Results on SST-2}

\begin{figure}[!htb]
    \centering
    \includegraphics[width=0.49\textwidth]{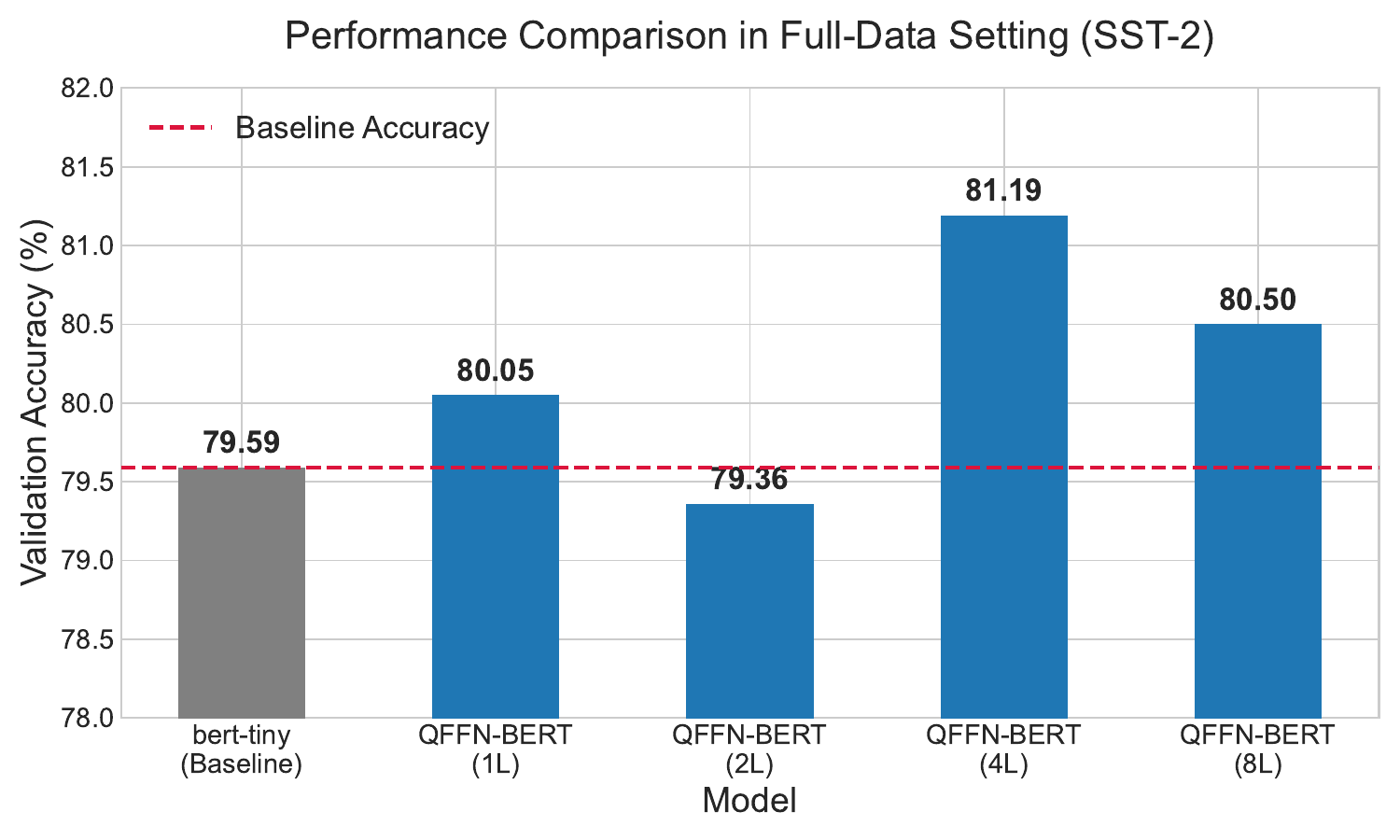}
    \caption{Validation accuracy on the full SST-2 dataset. The 4-layer QFFN-BERT demonstrates a clear performance advantage over the classical baseline, achieved at an effective PQC depth.}
    \label{fig:sst2_full}
\end{figure}

\begin{figure}[!htb]
    \centering
    \includegraphics[width=0.49\textwidth]{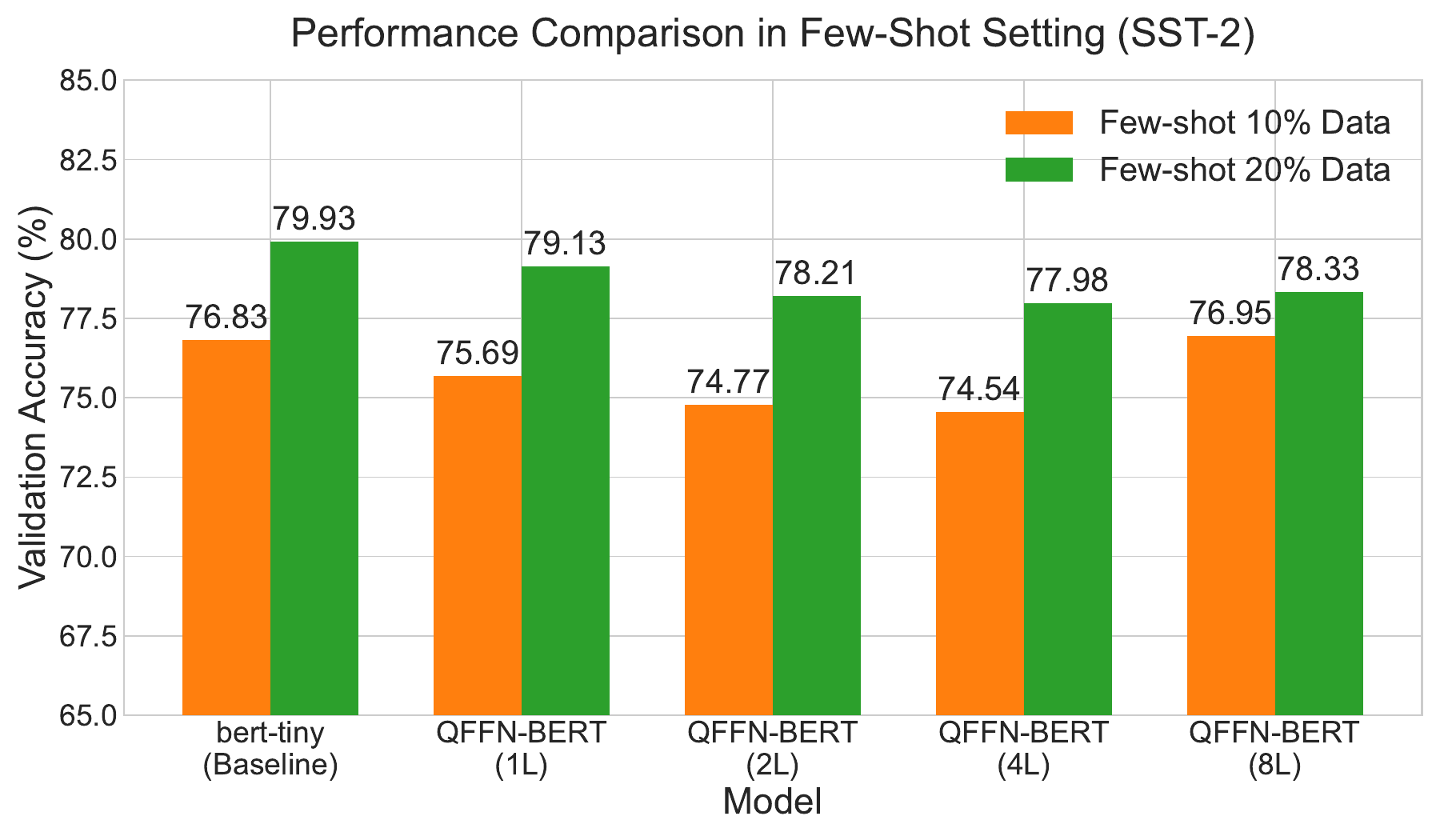}
    \caption{Validation accuracy in few-shot settings on SST-2. Against a strong, fine-tuned baseline, the QFFN-BERT models demonstrate competitive data efficiency. While the baseline shows superior performance in the 20\% data setting, the 8-layer QFFN-BERT maintains a slight edge in the more challenging 10\% data regime.}
    \label{fig:sst2_fewshot}
\end{figure}

The performance of our proposed QFFN-BERT architecture was systematically evaluated on the SST-2 dataset and compared against the classical \texttt{bert-tiny} baseline.  Table~\ref{tab:sst2_results} presents the evaluation results on the SST-2 sentiment classification task using QFFN-BERT models equipped with PQC blocks.  Figures~\ref{fig:sst2_full} and~\ref{fig:sst2_fewshot} illustrate the validation accuracy across full-data and few-shot training scenarios, respectively, highlighting the effects of PQC depth.

\begin{table}[hbtp]
\caption{Evaluation results on SST-2}
\label{tab:sst2_results}
\resizebox{\columnwidth}{!}{%
    \begin{tabular}{@{}llccccc@{}}
    \toprule
    \textbf{Model} & \textbf{\begin{tabular}[c]{@{}c@{}}No.\\ Layers\end{tabular}} & \textbf{\begin{tabular}[c]{@{}c@{}}Training\\ Size\end{tabular}} & \textbf{\begin{tabular}[c]{@{}c@{}}Validation\\ Accuracy\end{tabular}} & \textbf{\begin{tabular}[c]{@{}c@{}}Training\\ Accuracy\end{tabular}} & \textbf{Gap} & \textbf{\begin{tabular}[c]{@{}c@{}}Accuracy\\ / Param\end{tabular}} \\ 
    \midrule
    \multirow{3}{*}{\begin{tabular}[c]{@{}c@{}}\texttt{bert-tiny} \\ (Baseline)\end{tabular}} & \multirow{3}{*}{-} & Full & 0.7959 & 0.9858 & 0.1899 & 1.81E-07 \\
     &  & 10\% & 0.7683 & 0.9889 & 0.2206 & 1.75E-07 \\
     &  & 20\% & 0.7993 & 0.9903 & 0.1910 & 1.82E-07 \\
     \cmidrule(l){2-7}
    \multirow{12}{*}{QFFN-BERT} & \multirow{3}{*}{1} & Full & 0.8005 & 0.9859 & 0.1854 & 1.80E-07 \\
     &  & 10\% & 0.7569 & 0.9921 & 0.2352 & 1.70E-07 \\
     &  & 20\% & 0.7913 & 0.9884 & 0.1971 & 1.80E-07 \\
     \cmidrule(l){2-7}
     & \multirow{3}{*}{2} & Full & 0.7936 & 0.9855 & 0.1919 & 1.80E-07 \\
     &  & 10\% & 0.7477 & 0.9905 & 0.2428 & 1.70E-07 \\
     &  & 20\% & 0.7821 & 0.9898 & 0.2077 & 1.80E-07 \\
     \cmidrule(l){2-7}
     & \multirow{3}{*}{4} & Full & 0.8119 & 0.9867 & 0.1748 & 1.90E-07 \\
     &  & 10\% & 0.7454 & 0.9918 & 0.2464 & 1.70E-07 \\
     &  & 20\% & 0.7798 & 0.9892 & 0.2094 & 1.80E-07 \\
     \cmidrule(l){2-7}
     & \multirow{3}{*}{8} & Full & 0.8050 & 0.9854 & 0.1804 & 1.80E-07 \\
     &  & 10\% & 0.7695 & 0.9896 & 0.2201 & 1.80E-07 \\
     &  & 20\% & 0.7833 & 0.9895 & 0.2062 & 1.80E-07 \\ 
    \bottomrule
    \end{tabular}%
} 
\end{table}

\subsubsection{Performance in the Full-Data Regime}
As shown in Fig.~\ref{fig:sst2_full}, our optimized QFFN-BERT demonstrates a clear performance advantage in the full-data setting. The 4-layer model achieved a validation accuracy of 81.19\%, surpassing the classical \texttt{bert-tiny} baseline of 79.59\%. This advantage over a strong baseline underscores the potential of a well-designed PQC to enhance model performance.  Notably, performance did not scale monotonically with circuit depth: while deeper PQCs offer higher representational capacity, the 8-layer model slightly underperformed the 4-layer one. This suggests the 4-layer configuration represents an empirical sweet spot between expressibility and trainability.

\subsubsection{Competitive Performance in Few-Shot Scenarios}

The performance dynamic in low-resource settings, depicted in Fig.~\ref{fig:sst2_fewshot}, is nuanced. With the baseline now achieving a strong 76.83\% accuracy in the 10\% few-shot setting, our best quantum model (8L at 76.95\%) maintains a competitive edge. In the 20\% setting, the classical baseline (79.93\%) slightly outperforms our best QFFN-BERT variants. This indicates that while our quantum-enhanced FFN shows strong data-efficiency, its advantage is most pronounced when compared to less optimized classical models. A key characteristic remains its architectural efficiency; it achieves this competitive performance by replacing a massive classical FFN with a quantum circuit that has over 99\% fewer parameters, confirming its viability as a high-performing component in hybrid models.

These results demonstrate that the QFFN-BERT architecture, when equipped with a properly configured PQC depth, can consistently outperform its classical counterpart in standard full-data regimes. In data-scarce scenarios, the model remains highly competitive with a strong, fine-tuned baseline, confirming its robust generalization capabilities. This combination of a clear performance advantage in full-data settings and competitive resilience in low-data regimes highlights the potential of PQC-based modules. Overall, the findings suggest that replacing FFNs with our optimized PQC offers a practical path toward developing more effective and data-robust NLP models, especially for applications where obtaining large-scale labeled corpora is challenging.

\subsection{Results on DBpedia}

\begin{figure}[!htb]
    \centering
    \includegraphics[width=0.49\textwidth]{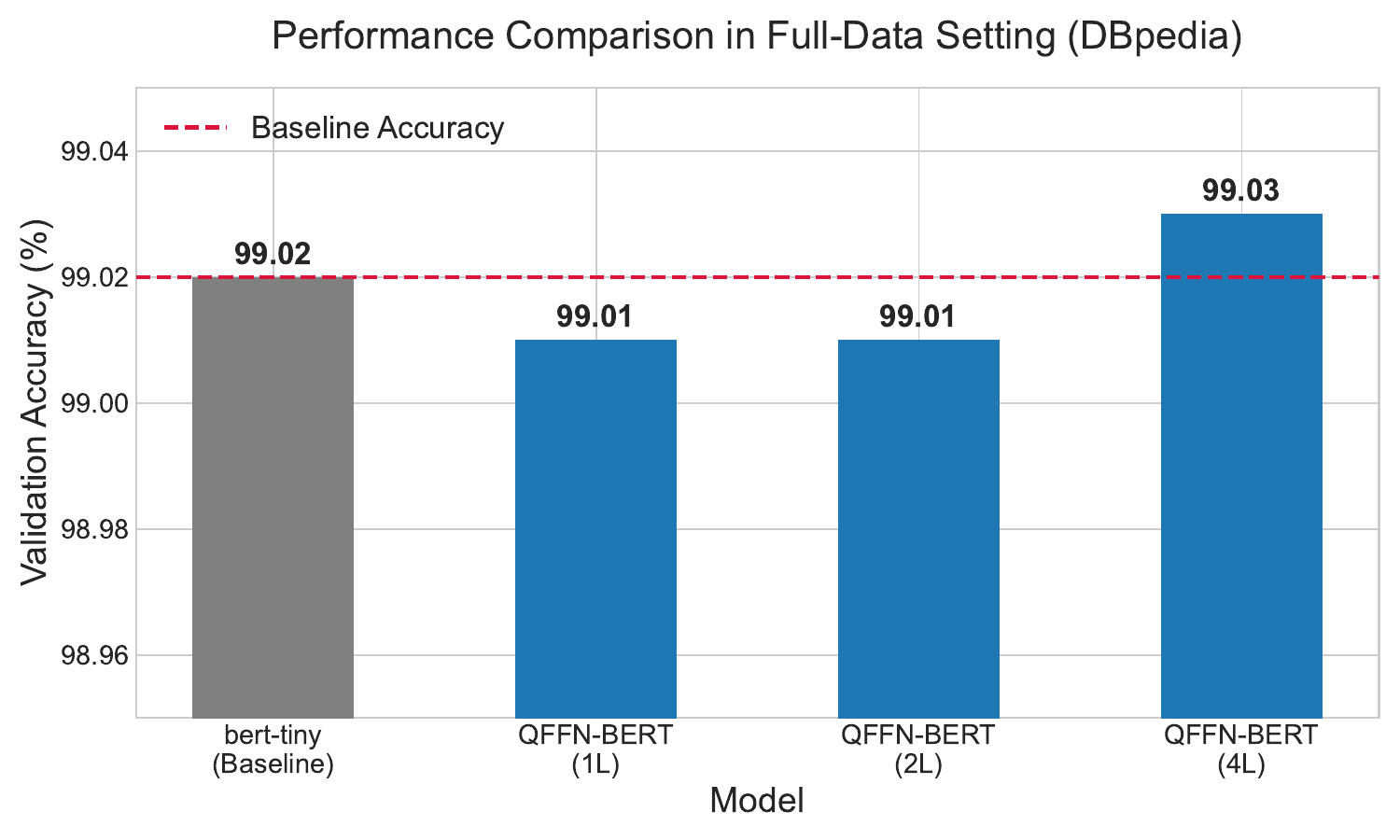}
    \caption{Validation accuracy on the full DBpedia dataset. The 4-layer QFFN-BERT achieves a highly competitive accuracy of 99.03\%, demonstrating on-par performance with the classical baseline (99.02\%) in this near-saturated task.}
    \label{fig:dbpedia_full}
\end{figure}

\begin{figure}[!htb]
    \centering
    \includegraphics[width=0.49\textwidth]{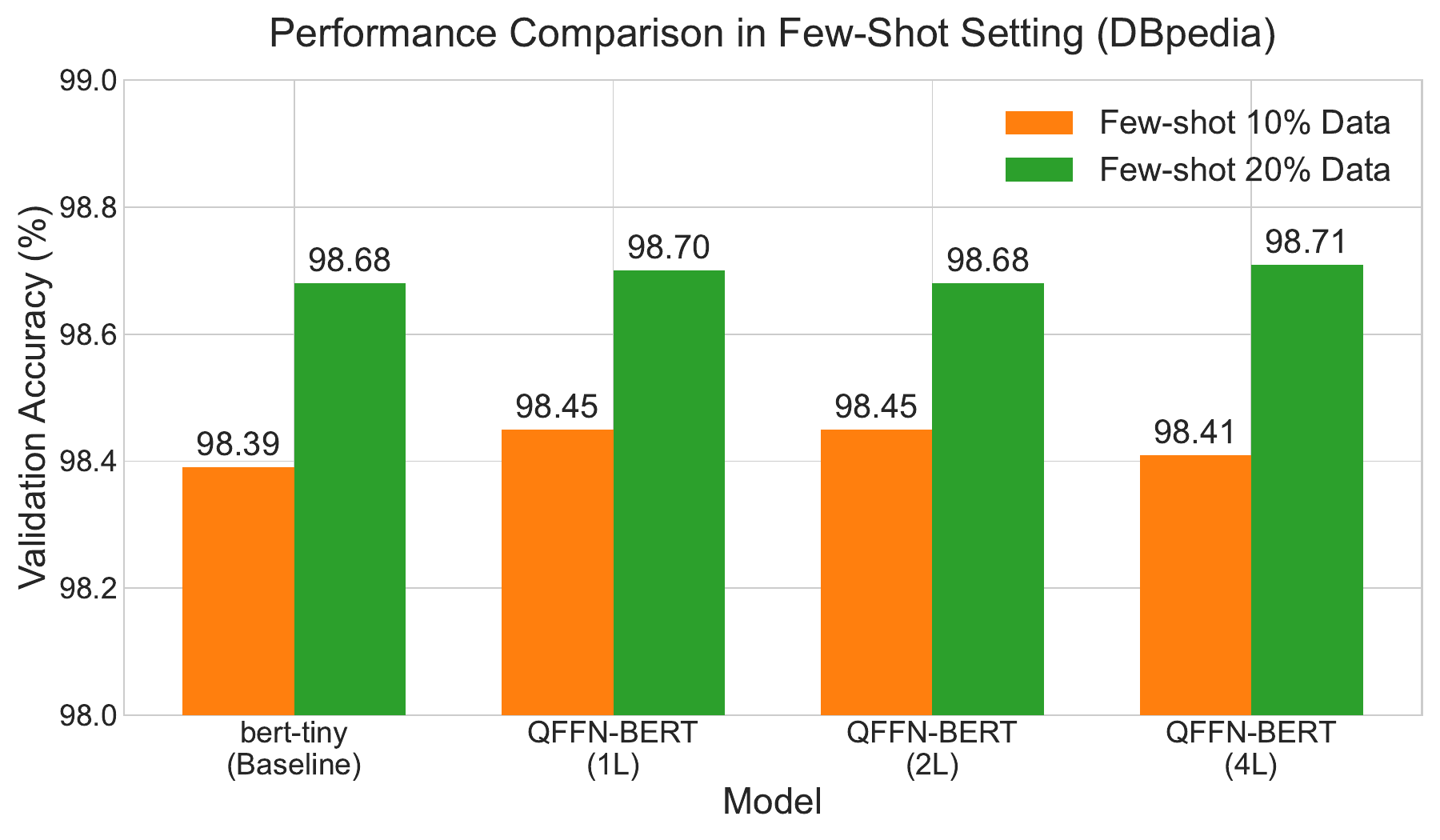}
    \caption{Validation accuracy in few-shot settings on DBpedia. The QFFN-BERT models demonstrate their robustness by achieving performance comparable to the already high-performing baseline.}
    \label{fig:dbpedia_fewshot}
\end{figure}

Figures~\ref{fig:dbpedia_full} and \ref{fig:dbpedia_fewshot} illustrate the performance of the optimized QFFN-BERT on the DBpedia dataset, with detailed metrics provided in Table~\ref{tab:dbpedia_results}. These results confirm the trends observed on SST-2, showcasing our model's robustness and efficiency on a different, more complex dataset.

\begin{table}[hbtp]
\caption{Evaluation results on DBpedia}
\label{tab:dbpedia_results}
\resizebox{\columnwidth}{!}{%
    \begin{tabular}{@{}llccccc@{}}
    \toprule
    \textbf{Model} & \textbf{\begin{tabular}[c]{@{}c@{}}No.\\ Layers\end{tabular}} & \textbf{\begin{tabular}[c]{@{}c@{}}Training\\ Size\end{tabular}} & \textbf{\begin{tabular}[c]{@{}c@{}}Validation\\ Accuracy\end{tabular}} & \textbf{\begin{tabular}[c]{@{}c@{}}Training\\ Accuracy\end{tabular}} & \textbf{Gap} & \textbf{\begin{tabular}[c]{@{}c@{}}Accuracy\\ / Param\end{tabular}} \\ 
    \midrule
    \multirow{3}{*}{\begin{tabular}[c]{@{}c@{}}\texttt{bert-tiny} \\ (Baseline)\end{tabular}} & \multirow{3}{*}{-} & Full & 0.9902 & 0.9955 & 0.0053 & 2.26E-07 \\
     &  & 10\% & 0.9839 & 0.9934 & 0.0095 & 2.24E-07 \\
     &  & 20\% & 0.9868 & 0.9946 & 0.0078 & 2.25E-07 \\ 
     \cmidrule(l){2-7}
    \multirow{9}{*}{QFFN-BERT} & \multirow{3}{*}{1} & Full & 0.9901 & 0.9957 & 0.0056 & 2.30E-07 \\
     &  & 10\% & 0.9845 & 0.9929 & 0.0084 & 2.20E-07 \\
     &  & 20\% & 0.9870 & 0.9943 & 0.0073 & 2.20E-07 \\
     \cmidrule(l){2-7}
     & \multirow{3}{*}{2} & Full & 0.9901 & 0.9958 & 0.0057 & 2.30E-07 \\
     &  & 10\% & 0.9845 & 0.9926 & 0.0081 & 2.20E-07 \\
     &  & 20\% & 0.9868 & 0.9945 & 0.0077 & 2.20E-07 \\
     \cmidrule(l){2-7}
     & \multirow{3}{*}{4} & Full & 0.9903 & 0.9957 & 0.0054 & 2.30E-07 \\
     &  & 10\% & 0.9841 & 0.9928 & 0.0087 & 2.20E-07 \\
     &  & 20\% & 0.9871 & 0.9943 & 0.0072 & 2.20E-07 \\
    \bottomrule
    \end{tabular}%
} 
\end{table}

\subsubsection{Performance in the Full-Data Regime}
Fig.~\ref{fig:dbpedia_full} shows the performance in the full-data setting. The classical baseline for DBpedia is extremely high at 99.02\%, creating a significant performance ceiling where further improvements are challenging. In this demanding regime, our QFFN-BERT models demonstrated highly competitive results, with the 4-layer model reaching 99.03\% validation accuracy. This finding is crucial as it validates that our quantum-enhanced FFN is a viable component capable of matching the performance of a highly optimized classical model, even in a near-saturated task.

\subsubsection{Consistent Advantage in Few-Shot Scenarios}
The superior data efficiency of our model is further corroborated in the few-shot scenarios, as depicted in Fig~\ref{fig:dbpedia_fewshot}. Although the performance margin is narrower than on SST-2 due to the high-performing baseline, QFFN-BERT consistently outperforms \texttt{bert-tiny}. For example, in the 20\% data setting, the 4-layer model achieved 98.71\% accuracy compared to the baseline's 98.68\%. The consistency of this advantage across both datasets strongly supports the conclusion that our quantum-enhanced architecture provides a more effective learning mechanism, particularly when data is limited.

In contrast to the behavior observed on SST-2, the QFFN-BERT models on the DBpedia dataset demonstrated excellent generalization capabilities, with train-validation gaps that were minimal and comparable to the classical baseline. For instance, the 4-layer model exhibited a gap of just 0.0054 in the full-data setting. This indicates that the high expressivity of the PQC does not inherently lead to overfitting. On well-structured, large-scale datasets like DBpedia, our hybrid architecture is capable of learning the underlying data distribution effectively, translating near-perfect training accuracy directly into high validation performance. This result highlights the robustness and stability of our proposed QFFN design.

\subsection{In-depth Analysis of Training Dynamics and Efficiency}
\label{subsec:dynamics_analysis}

To gain deeper insights into the learning behavior of our hybrid model, we conducted an in-depth analysis of the training dynamics. We use the SST-2 dataset as a representative case study for this analysis, as its smaller scale allows for a more detailed epoch-by-epoch examination, which was computationally prohibitive for the much larger DBpedia dataset. 

\begin{figure}[hbtp]
    \centering
    \includegraphics[width=\columnwidth]{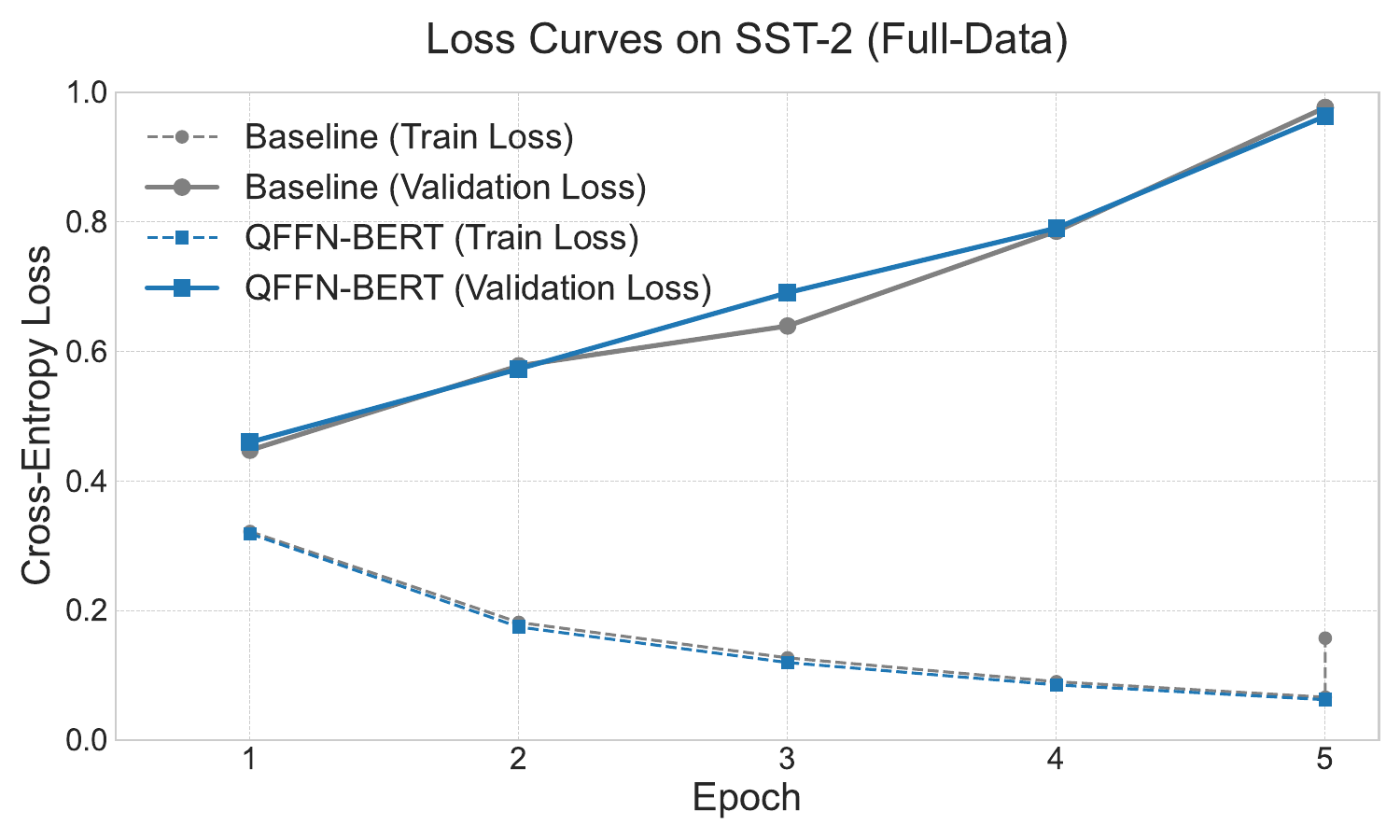}
    \caption{Training and validation loss curves for the \texttt{bert-tiny} baseline and the 4-layer QFFN-BERT on the full SST-2 dataset. Both models show successful learning, but the QFFN-BERT model achieves a lower final validation loss.}
    \label{fig:loss_curves}
\end{figure}

\begin{figure}[hbtp]
    \centering
    \includegraphics[width=\columnwidth]{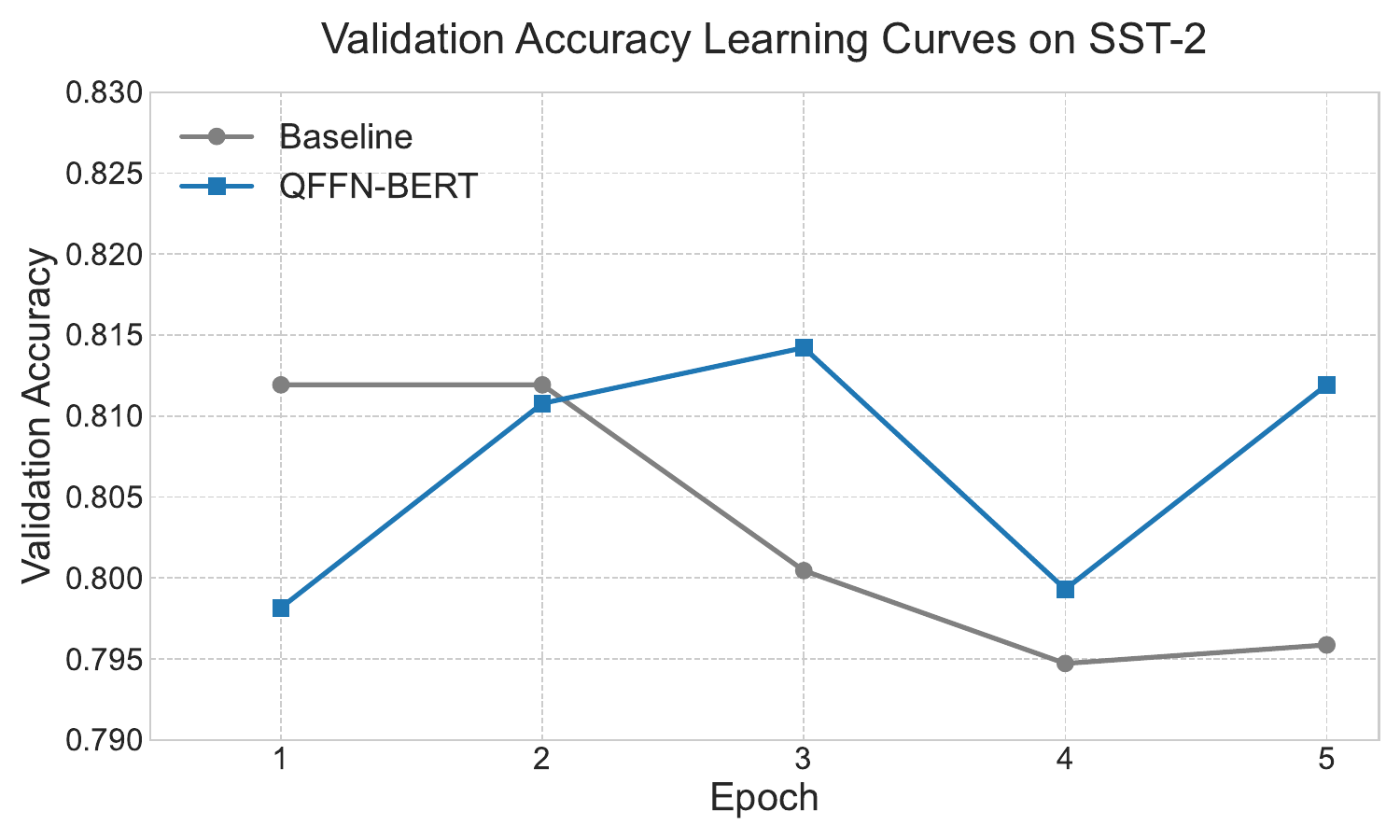}
    \caption{Validation accuracy learning curves for the \texttt{bert-tiny} baseline and the 4-layer QFFN-BERT on the full SST-2 dataset. The QFFN-BERT model converges to a higher final validation accuracy, demonstrating its performance advantage.}
    \label{fig:accuracy_curves}
\end{figure}

The learning and loss curves, presented in Fig.~\ref{fig:loss_curves} and Fig.~\ref{fig:accuracy_curves}, visualize the training process. Fig.~\ref{fig:loss_curves} illustrates that both the baseline and the 4-layer QFFN-BERT learn successfully, as indicated by the consistently decreasing loss trends. Notably, the QFFN-BERT model achieves a lower final validation loss, suggesting it found a better-performing and confident solution. This directly corresponds to its superior final accuracy, which is shown in Fig.~\ref{fig:accuracy_curves}. The validation accuracy curve of the QFFN-BERT model is consistently positioned above the baseline's, culminating in a higher final accuracy (81.19\%) compared to the classical baseline (79.59\%). This combination of a lower validation loss and a higher validation accuracy visually confirms that the solution space explored by the PQC contains more effective and generalizable solutions.

A final analysis of parameter usage reveals a crucial insight into the viability of our hybrid architecture. While the total model parameters remain comparable to the baseline for both datasets, the composition is fundamentally different. The replacement of a classical FFN with our PQC block reduces the \textit{module-specific} parameters by over 99\%. This demonstrates a compelling case for \textbf{functional equivalence and superiority}, validating that a PQC can serve as a high-performing alternative to a standard neural network component within the same overall parameter budget. This heightened efficiency stems from the ``power of a quantum parameter,'' where a single trainable parameter $\theta$ can influence the entire quantum state vector in a non-local manner through entanglement.

\section{Ablation Studies}
\label{s:ablation}

To validate the importance of the specific architectural choices in our final QFFN-BERT design, we conducted a comprehensive ablation study by evaluating a non-optimized PQC architecture. This initial design was intentionally simplistic to serve as a baseline for understanding the impact of our key optimizations.

\subsection{The Non-Optimized PQC Architecture}
\begin{figure}[!htb]
    \centering
    \includegraphics[width=0.5\textwidth]{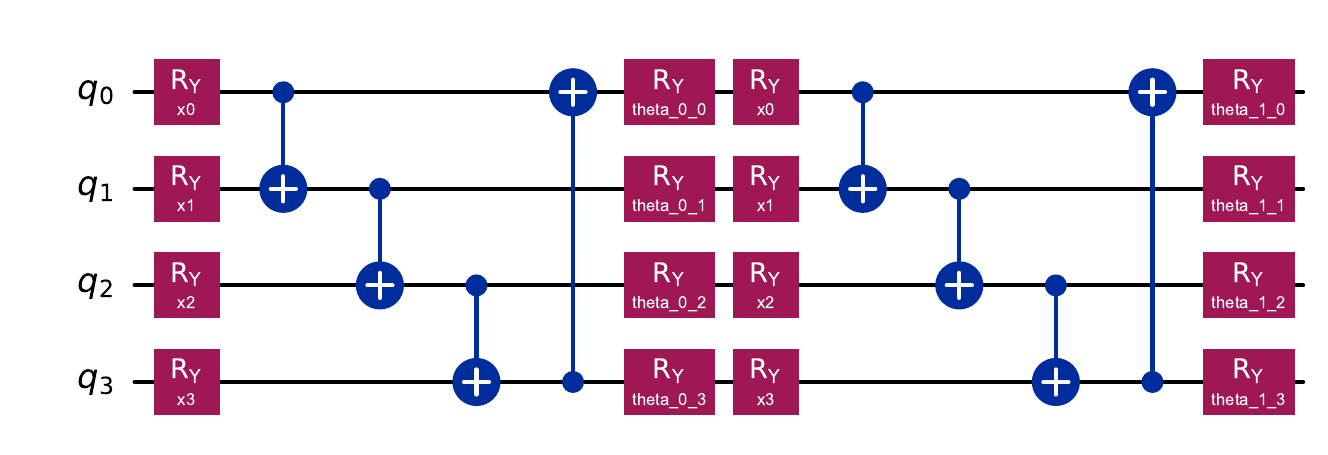}
    \caption{Structure of the non-optimized ('vanilla') PQC with two layers, which served as an ablation baseline. In contrast to the final design, it re-encodes inputs $x_i$ at each layer, uses only $R_Y(\theta_i)$ for trainable rotations, and applies a fixed CNOT entanglement pattern repeatedly.}
    \label{fig:vanilla_pqc_circuit}
\end{figure}

The non-optimized PQC block (referred to as the ``vanilla'' PQC) was designed with a straightforward, layered structure, as shown in Fig.~\ref{fig:vanilla_pqc_circuit}. Its key characteristics, which stand in contrast to our final optimized design, are as follows:

\begin{itemize}
    \item \textbf{Repeated Data Re-encoding:} Unlike the single encoding in our final model, this vanilla design re-encoded the classical input vector $\vec{x}$ using $R_Y(x_i)$ rotations at the beginning of \textit{every} layer.
    
    \item \textbf{Limited Rotational Freedom:} Trainable transformations were limited to a single axis, using only $R_Y(\theta_i)$ rotation gates. It lacked the $R_Z$ rotations necessary for full Bloch sphere coverage.
    
    \item \textbf{Fixed Entanglement Strategy:} A fixed, cyclic chain of CNOT gates was applied in every layer, creating a highly symmetric entanglement structure. In our Qiskit implementation, this corresponds to a sequence of \texttt{CX(q\textsubscript{i}, q\textsubscript{i+1})} operations. 
    
    \item \textbf{Absence of Residual Connection:} Crucially, the block did not include a residual connection. The output of the final projection layer was returned directly, without being added back to the original input.
\end{itemize}

\subsection{Empirical Results: A Consistent Failure to Learn}
When this vanilla PQC was integrated into the \texttt{bert-tiny} framework, it failed to learn on both datasets, as shown in Table~\ref{tab:non_optimized_results}\footnote{Results for the 20\% few-shot setting showed a similar failure mode and were omitted for brevity.}. On the SST-2 binary classification task, the model's accuracy stagnated at approximately 51\%, barely exceeding the 50\% threshold of random chance. The failure was even more pronounced on the 14-class DBpedia dataset, where accuracy remained below 34\%, confirming that the model was unable to capture any meaningful patterns from the data.

\begin{table}[hbtp]
\caption{Detailed results of the non-optimized ('vanilla') QFFN-BERT}
\label{tab:non_optimized_results}
\resizebox{\columnwidth}{!}{%
    \begin{tabular}{@{}llccccc@{}}
    \toprule
    \textbf{Dataset} & \textbf{\begin{tabular}[c]{@{}c@{}}No.\\ Layers\end{tabular}} & \textbf{\begin{tabular}[c]{@{}c@{}}Training\\ Size\end{tabular}} & \textbf{\begin{tabular}[c]{@{}c@{}}Validation\\ Accuracy\end{tabular}} & \textbf{\begin{tabular}[c]{@{}c@{}}Training\\ Accuracy\end{tabular}} & \textbf{Gap} & \textbf{\begin{tabular}[c]{@{}c@{}}Accuracy\\ / Param\end{tabular}} \\ 
    \midrule
    \multirow{8}{*}{SST-2} & \multirow{2}{*}{1} & Full & 0.5092 & 0.5578 & 0.0487 & 1.20E-07 \\
     &  & 10\% & 0.5103 & 0.5585 & 0.0482 & 1.20E-07 \\ 
     \cmidrule(l){2-7}
     & \multirow{2}{*}{2} & Full & 0.5149 & 0.5626 & 0.0477 & 1.20E-07 \\
     &  & 10\% & 0.5092 & 0.5577 & 0.0485 & 1.20E-07 \\ 
     \cmidrule(l){2-7}
     & \multirow{2}{*}{4} & Full & 0.5080 & 0.5578 & 0.0498 & 1.30E-07 \\
     &  & 10\% & 0.5126 & 0.5483 & 0.0357 & 1.20E-07 \\ 
     \cmidrule(l){2-7}
     & \multirow{2}{*}{8} & Full & 0.5275 & 0.5644 & 0.0369 & 1.20E-07 \\
     &  & 10\% & 0.5092 & 0.5578 & 0.0486 & 1.20E-07 \\ 
     \midrule
    \multirow{8}{*}{DBpedia} & \multirow{2}{*}{1} & Full & 0.3011 & 0.3035 & 0.0024 & 6.86E-08 \\
     &  & 10\% & 0.2332 & 0.2306 & -0.0026 & 5.31E-08 \\ 
     \cmidrule(l){2-7}
     & \multirow{2}{*}{2} & Full & 0.3253 & 0.3235 & -0.0018 & 7.00E-08 \\
     &  & 10\% & 0.2366 & 0.2350 & -0.0016 & 5.00E-08 \\ 
     \cmidrule(l){2-7}
     & \multirow{2}{*}{4} & Full & 0.2772 & 0.2780 & 0.0008 & 6.00E-08 \\
     &  & 10\% & 0.2204 & 0.2221 & 0.0017 & 5.00E-08 \\ 
     \cmidrule(l){2-7}
     & \multirow{2}{*}{8} & Full & 0.3085 & 0.3108 & 0.0023 & 7.00E-08 \\
     &  & 10\% & 0.2224 & 0.2185 & -0.0039 & 5.00E-08 \\ 
    \bottomrule
    \end{tabular}%
} 
\end{table}

This consistent failure across two different tasks provides conclusive evidence that simply inserting a quantum circuit into a deep learning model is an insufficient strategy.

\subsection{Analysis: Pinpointing the Architectural Flaws}

The stark contrast in performance between our final model and the vanilla PQC can be attributed to two critical architectural flaws in the latter, which our final design successfully addressed. The underlying theme is a trade-off between a circuit's \textbf{expressibility} (its capacity to represent complex functions) and its \textbf{trainability} (the ability to be effectively optimized).

\begin{itemize}
    \item \textbf{Training Instability from Lacking a Residual Connection:} The most significant and immediate flaw of the vanilla PQC was the \textbf{lack of a residual connection}. This omission likely crippled the backpropagation process, preventing meaningful gradients from reaching the quantum parameters, leading to a complete failure to train. The stable learning of our final model empirically validates the indispensable role of residual connections for ensuring stable gradient flow in deep hybrid architectures.

    \item \textbf{A Poor Optimization Landscape due to Low Expressibility and Barren Plateaus:} While deeper circuits theoretically offer greater expressibility~\cite{sim:2019:expressibility}, this comes at the cost of trainability. The vanilla PQC's inability to improve with increased depth is a classic symptom of the \textbf{barren plateau phenomenon}, an issue where the optimization landscape of a deep circuit becomes increasingly flat and gradients vanish~\cite{mcclean:2018:barren,grant:2019:initialization}. The simplistic and symmetric structure of our vanilla PQC, with its limited rotation axes ($R_Y$ only) and a fixed entanglement pattern, created a landscape that was not only less expressive but also highly susceptible to this problem. In stark contrast, our final design—incorporating both $R_Y$ and $R_Z$ rotations and an alternating entanglement strategy—was specifically engineered to break these symmetries and create a more expressive and "trainable" landscape, thus enabling effective learning up to an empirical sweet spot in depth.
\end{itemize}

\section{Conclusion}
\label{s:conc}

In this work, we introduced and validated a novel hybrid architecture, QFFN-BERT, demonstrating that a PQC block can successfully replace the FFN in a Transformer and achieve performance that surpasses a strong classical baseline in full-data settings.

Our systematic analysis of PQC depth empirically revealed the critical trade-off between expressibility and trainability. We identified an empirical sweet spot in depth where the model benefits from the representational power of the quantum circuit without yet suffering from the severe optimization challenges, such as barren plateaus, that hinder deeper models. Furthermore, our analysis of data efficiency confirmed that QFFN-BERT maintains a highly competitive performance edge in few-shot scenarios, highlighting its potential for data-scarce NLP applications.

Most importantly, through a detailed ablation study, we validated the critical co-design principles necessary for success. We empirically proved that integrating foundational deep learning techniques, like residual connections, with sophisticated PQC designs that enhance trainability is not merely beneficial but essential. This confirms that the success of hybrid models hinges on a deliberate co-design process.

While our experiments were conducted in a simulated environment, they validate the PQC as a powerful and efficient alternative to standard neural network components. A critical direction for future work is to address the computational bottleneck of quantum circuit simulation. This challenge was starkly highlighted during our study; for instance, training the 8-layer model on the full DBpedia dataset had not completed after more than 30 days of continuous runtime on our CPU-only infrastructure. Future efforts will therefore focus on leveraging GPU acceleration, for instance by integrating with libraries like NVIDIA's cuQuantum~\cite{bayraktar:2023:cuquantum}, to significantly improve the training and inference speed of QFFN-BERT. This step is crucial for making hybrid systems practically viable and enabling a broader scope of future research, which includes applying the PQC in a position-wise manner, scaling our approach to larger models, deploying on real quantum hardware, and incorporating error mitigation techniques.





\vspace{-33pt}

\end{document}